\documentclass[letterpaper, 10 pt, conference]{ieeeconf}  

\IEEEoverridecommandlockouts                              

\usepackage{graphics}
\usepackage{amsmath}
\usepackage{amssymb}
\usepackage[colorlinks, linkcolor=blue, citecolor=blue]{hyperref}
\usepackage{cite}
\usepackage{algorithm}
\usepackage{url}
\usepackage{algpseudocode}
\usepackage{epsfig}
\usepackage{graphicx}
\usepackage{caption}
\usepackage{subfigure}
\usepackage{mathrsfs}
\usepackage{array}
\usepackage{multirow}
\usepackage{multicol}
\usepackage{booktabs}
\usepackage{makecell}
\usepackage{xcolor}
\usepackage{color}
\usepackage{amsfonts}
\usepackage{gensymb}
\usepackage{threeparttable}
\usepackage{textcomp}

\IEEEaftertitletext{\vspace{-1em}}

\title{\LARGE \bf
ETP-R1: Evolving Topological Planning with Reinforcement Fine-tuning for Vision-Language Navigation in Continuous Environments\\[-1.0em]
}

\author{
Shuhao Ye$^{1}$, Sitong Mao$^{2}$, Yuxiang Cui$^{3}$, Xuan Yu$^{1}$, Shichao Zhai$^{1}$, Wen Chen$^{2}$, Shunbo Zhou$^{2}$, Rong Xiong$^{1}$, \\
Yue Wang$^{1}$$^{\dagger}$
\thanks{$^{1}$Zhejiang University, Hangzhou, China}%
\thanks{$^{2}$Huawei Technologies Co., Ltd}%
\thanks{$^{3}$Zhejiang Humanoid Robot Innovation Center, Ningbo, China}%
\thanks{$\dag$Corresponding author: Yue Wang (email: wangyue@iipc.zju.edu.cn)
}
}

\begin{document}

\maketitle
\thispagestyle{empty}
\pagestyle{empty}

\begin{abstract}

Vision-Language Navigation in Continuous Environments (VLN-CE) requires an embodied agent to navigate towards target in continuous environments, following natural language instructions. While current graph-based methods offer an efficient, structured approach by abstracting the environment into a topological map and simplifying the action space to waypoint selection, they lag behind methods based on Large Vision-Language Models (LVLMs) in leveraging large-scale data and advanced training paradigms.
In this paper, we try to bridge this gap by introducing ETP-R1, a framework that applies the paradigm of scaling up data and Reinforcement Fine-Tuning (RFT) to a graph-based VLN-CE model. To build a strong foundation, we first construct a high-quality, large-scale pretraining dataset using the Gemini API. This dataset consists of diverse, low-hallucination instructions for topological trajectories, providing rich supervision for our graph-based policy to map language to topological paths. This foundation is further strengthened by unifying data from both R2R and RxR tasks for joint pretraining. Building on this, we introduce a three-stage training paradigm, which culminates in the first application of closed-loop, online RFT to a graph-based VLN-CE model, powered by the Group Relative Policy Optimization (GRPO) algorithm. Extensive experiments demonstrate that our approach is highly effective, establishing new state-of-the-art performance across all major metrics on both the R2R-CE and RxR-CE benchmarks.
Our code is available at \url{https://github.com/Cepillar/ETP-R1}.
\end{abstract}

\section{Introduction}

Vision-Language Navigation (VLN) \cite{VLN} requires embodied agents to reach a target location in unseen environments by following natural language instructions. To achieve robust perception and spatial reasoning, agents in this task are typically provided with panoramic RGBD observations, which grant a fine-grained, 3D understanding of the environment. While the original VLN task was defined for discrete environments, recent research has increasingly shifted their focus to the more realistic setting of VLN in Continuous Environments (VLN-CE) \cite{krantz2020beyond}, where agents operate in the Habitat simulator \cite{savva2019habitat} using motor actions to obtain observations in a continuous space.


Despite recent progress, existing approaches based on Large Vision-Language Models (LVLMs) \cite{wei2025streamvln, qi2025vlnr1} face key inefficiencies when applied to VLN-CE. First, their token-based input format is highly redundant: as the agent moves through overlapping views, token counts grow linearly, forcing a trade-off between retaining history and controlling computational cost. Second, hampered by the fact that LVLMs are rarely pretrained with extensive panoramic data, most approaches process only a forward-facing view, discarding the valuable panoramic context provided by the task itself. With no suitable high-level action space built upon this unstructured representation, the agent is confined to a low-level  action space (e.g., turn, move forward), requiring it to learn long-horizon navigation skills from trivial motor actions.

Some other approaches leverage graph representation \cite{an2024etpnav, an2022bevbert, hnr}, which is difficult for prior LVLM-based methods to handle. They construct a sparse topological map online using a pretrained waypoint predictor \cite{hong2022bridging}. This representation is both compact and informative: the agent only needs to decide which waypoint to go next, while a deterministic controller executes the low-level motion. By decomposing the problem into high-level waypoint selection and low-level execution, graph-based methods reduce end-to-end complexity and retain full panoramic structure.

\begin{figure}[t]
    \centering
    \begin{minipage}[t]{1\linewidth}
    \centering
    \includegraphics[width=1\linewidth]{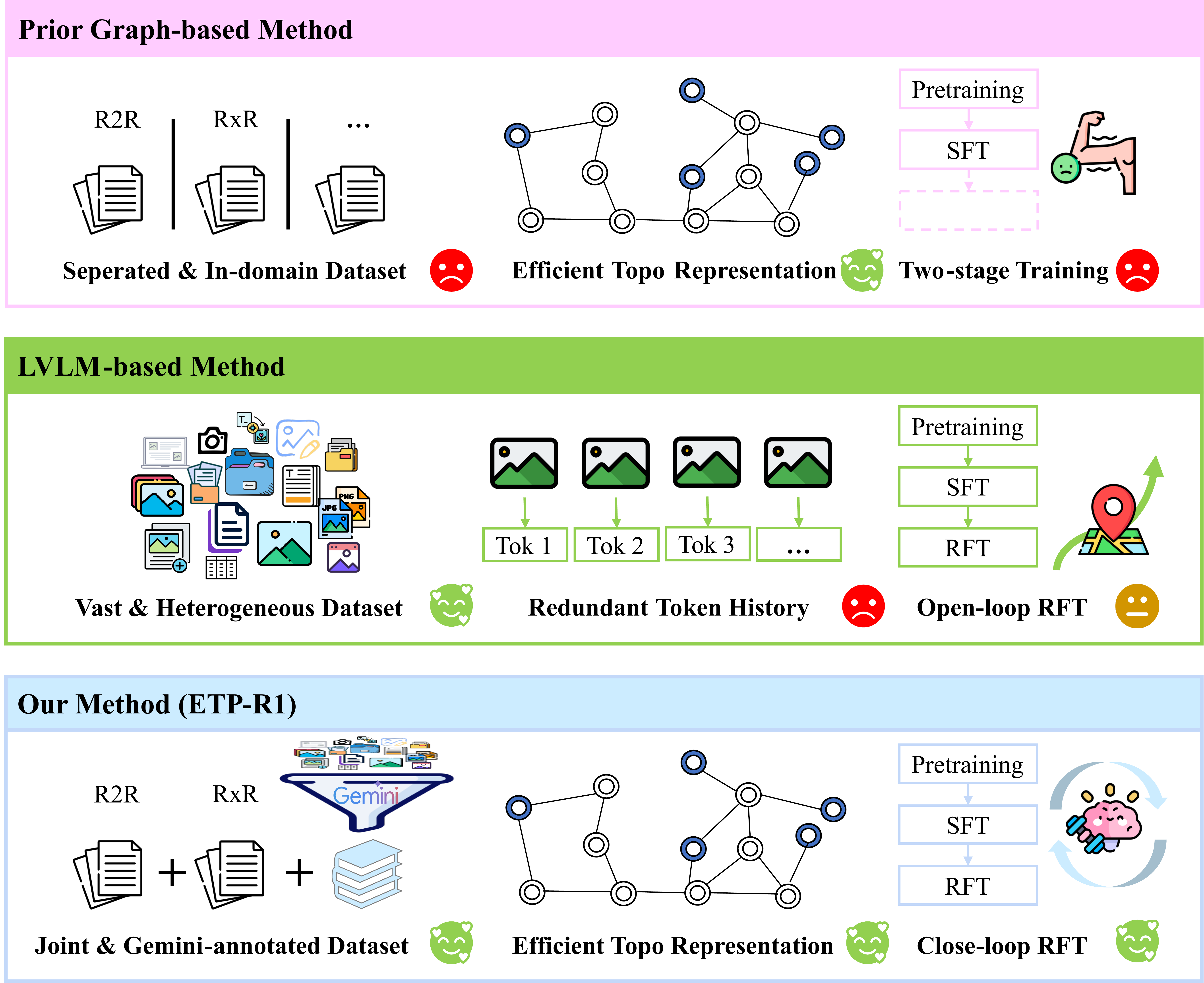}
    \end{minipage}
    \caption{\small{Comparison of three approaches' pipelines. Our work employs joint dataset pretraining and uses Gemini to generate new data; leverages efficient graph representations; and applies closed-loop RFT to graph-based VLN models for the first time.}}
    \label{fig:main}
    \vspace{-20pt}
\end{figure}

However, existing graph-based models have not fully capitalized on their advantages by embracing the large-scale training paradigms that powered LVLMs. The adaptability of LVLMs originating from their simple input format (raw images and text), which allows them to be trained on vast, heterogeneous datasets. In contrast, the structured, graph-based representation restricts these models to navigation-specific data. But even with this constraint, current graph-based approaches still overlook two valuable data sources: the rich linguistic knowledge from powerful LVLMs (which could enhance synthetic instructions) and the collective task data from multiple navigation datasets (e.g., R2R \cite{VLN} and RxR \cite{RXR}), which are typically used in isolation \cite{an2024etpnav, hnr, g3d} for task-specific training. Furthermore, these methods typically omit the reinforcement fine-tuning (RFT) stage—another missed opportunity, as the waypoint-based action space is well-suited for efficient, closed-loop RFT. 
For LVLM-based methods trained offline, RFT is just limited to refining their single-turn output in an open-loop fashion regarding the full trajectory, as the model cannot receive new environmental feedback. 
Graph-based methods, however, are well positioned for RFT that can iteratively optimize decisions along an entire trajectory.

Motivated by the analysis above, we introduce ETP-R1, a graph-based VLN framework that evolve topological planning with reinforcement fine-tuning. 
Our approach builds \textbf{an enhanced pretraining foundation} to enable \textbf{a novel online RFT paradigm} for graph-based VLN.
To establish a strong foundation for fine-tuning, We first re-annotate the Prevalent dataset \cite{hao2020towards} using the Gemini API \cite{team2023gemini} to enrich linguistic diversity and mitigate hallucinations. We then unify the pretraining of R2R and RxR into a single joint process, enabling our model to learn from a broader range of instructions. Our resulting 0.5-billion-parameter model is powered by a cross-modal planning network, featuring a Dual-Phase Fusion Transformer (DPFT) to effectively align language instructions with the graph representation. 
Building on this foundation, the agent is fine-tuned through two subsequent online stages. The process begins with an online Supervised Fine-Tuning (SFT) stage to adapt the model to the interactive environment. The final stage is online RFT, which—to the best of our knowledge—is first applied to graph-based VLN models. This stage, powered by the Group Relative Policy Optimization (GRPO) algorithm \cite{shao2024deepseekmath}, allows the agent to refine its navigation policy in a closed-loop manner by reinforcing more successful trajectories.




In summary, our main contributions are: 
\begin{itemize}
\item We propose a training-free instruction annotation pipeline using the Gemini API to generate a large-scale, high-quality pretraining dataset with rich linguistic diversity and fewer hallucinations.

\item We propose to jointly pretrain a single model on diverse VLN datasets (R2R and RxR), further enhancing its generalization capabilities.

\item We equip graph-based VLN with a three-stage training paradigm, featuring an online RFT stage for the first time, which is powered by closed-loop GRPO.

\item Through extensive experiments, our proposed framework establishes new SOTA on both the R2R-CE and RxR-CE benchmarks, validating the effectiveness of our large-scale training and fine-tuning strategies.
\end{itemize}

\section{Related Work}

\subsection{VLN in Continuous Environments (VLN-CE)}

To address VLN-CE task \cite{krantz2020beyond}, early approaches attempted to directly map visual and textual inputs to low-level actions \cite{law}, but their performance was often limited by inefficient action space. The introduction of the waypoint predictor \cite{krantz2021waypoint, hong2022bridging} marked a significant evolution, shifting the problem from low-level control to high-level waypoint selection. This waypoint-based paradigm was further formalized by graph-based methods \cite{an2024etpnav}, which organize predicted waypoints into a topological graph and employ a Transformer-based planner for high-level planning. While subsequent works within this framework \cite{an2022bevbert, hnr, g3d} have focused on improving front-end visual representations to aid planning, the performance gains from stacking more powerful feature extractors have begun to plateau. 

In parallel, the rapid advancement of Large Language Models has inspired a new line of research using LLMs/LVLMs as navigation agents \cite{wei2025streamvln, qi2025vlnr1}. The powerful pretrained weights of these models afford a deeper understanding of language instructions. However, current LVLM-based methods are typically limited in two ways: they have difficulty processing panoramic images, often relying on a monocular view, and they cannot effectively leverage topological structures, resorting to the history of image frames as their environmental representation.

Alongside novel architectures, the pretraining data is critical to performance. Many prior works trained "speaker" models to annotate instructions \cite{fried2018speaker, wang2023scalingvln, hao2020towards, wang2022marky}. However, this approach has notable drawbacks: the generated language is stylistically confined to the patterns of its training dataset, and the resulting instructions often suffer from hallucinations. In this work, we leverage a powerful, general-purpose LVLM (Gemini \cite{team2023gemini}) via its API to achieve instruction annotation with high linguistic diversity and a low rate of hallucination.

\subsection{Reinforcement learning (RL) in VLN-CE}

While RL was a common technique in discrete VLN \cite{chen2022reinforced, wang2018look}, its direct application to VLN-CE is expensive due to the expanded action space and longer trajectory length. Krantz et al. \cite{krantz2021waypoint} first apply RL to the waypoint predictor. Although this reduced the learning complexity, the agent still faced the difficult dual task of generating waypoints that were both navigable and instruction-compliant. Recently, the graph-based VLN \cite{an2024etpnav} further decouples waypoint generation from instruction following, which is an ideal framework for implementing RL.

Concurrently, the training paradigms of LVLMs \cite{shao2024deepseekmath} have shown that SFT followed by RL—a process known as RFT—is a highly efficient approach. Qi et al. \cite{qi2025vlnr1} introduced RFT to a LVLM-based agent. However, as their approach is based on offline training, the RFT is limited to fine-tuning a single-turn language output. This renders their RFT process actually open-loop with respect to the interactive navigation task. In contrast, our work introduces the first closed-loop RFT paradigm for the efficient graph-based VLN framework.

\begin{figure*}[h]
    \centering
    \begin{minipage}[t]{1\linewidth}
    \centering
    \includegraphics[width=1\linewidth]{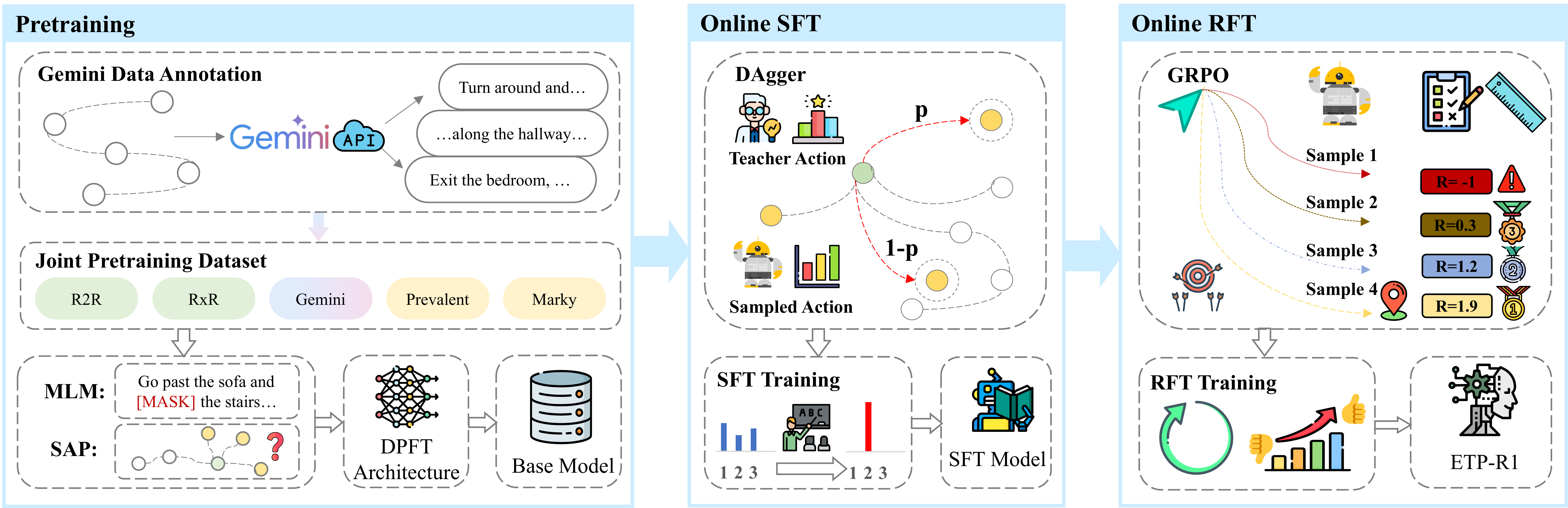}
    \end{minipage}
    \caption{\small{Overview of our approach. Our work focuses on pretraining and online RFT stages within a three-stage training paradigm.}}
    \label{fig:overview}
    \vspace{-15pt}
\end{figure*}

\section{Method}
\textit{Task setup: } In the VLN-CE task, an agent is initialized at a start location within an unseen environment and given a natural language instruction $W=\{w_i\}_{i=1}^{N_t}$. At each timestep, the agent receives a panoramic observation $O_t$ composed of 12 RGBD image pairs $\{I^{rgb}_k, I^{d}_k\}_{k=1}^{12}$, which evenly sample a 360° horizontal view. Guided by the instruction and current observation, the agent navigates towards the target by sequentially taking actions from four choices: MOVE FORWARD (0.25m), TURN LEFT/RIGHT (15°), and STOP. An episode is considered successful if the agent executes the STOP action within a predefined distance threshold of the target location before reaching a maximum step limit.

\textit{Overview of Our Approach: } We adopt the established topological mapping and low-level control strategies from ETPNav \cite{an2024etpnav}, the details of which are omitted for brevity. As shown in Figure \ref{fig:overview}, our work is centered around the pretraining and online RFT stages within a three-stage training paradigm. Our primary innovations are detailed in the subsequent sections: Section \ref{sec:annotation} introduces our training-free pretraining data annotation pipeline using Gemini; Section \ref{sec:model} presents our cross-modal planning network; and Section \ref{sec:train} details our three-stage training paradigm, emphasizing our unified R2R+RxR pretraining and the closed-loop RFT.

\begin{figure*}[h]
    \centering
    \begin{minipage}[t]{1\linewidth}
    \centering
    \includegraphics[width=\linewidth]{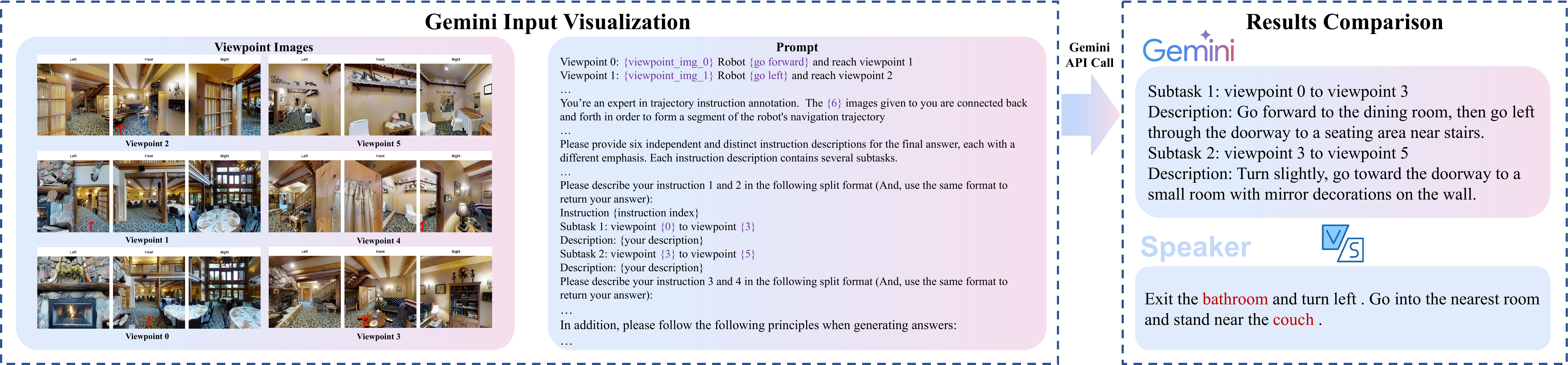}
    \end{minipage}
    \caption{\small{Visualization of the Gemini instruction annotation. The original Prevalent speaker annotation result is also included for comparison.}}
    \label{fig:data}
    \vspace{-15pt}
\end{figure*}

\subsection{Training-free Pretraining Data Annotation} 
\label{sec:annotation}
To enhance the quality and linguistic diversity, we re-annotate the instruction of Prevalent \cite{hao2020towards} dataset using Gemini 2.0 Flash API \cite{team2023gemini}. Our prompt, which is fed to Gemini, consists of carefully designed visual and textual components.

We represent the visual trajectory as an ordered sequence of viewpoint images. Each viewpoint in the sequence corresponds to a specific node from the predefined graph \cite{VLN}. To avoid processing panoramic images while still providing sufficient field of view, we create a composite image to represent each viewpoint. Specifically, we horizontally stitch the left, front, and right monocular views (90° FOV for each image) in order. Furthermore, to explicitly highlight the direction of travel, we superimpose a red arrow on the image pointing towards the next viewpoint. Fig.\ref{fig:data} shows examples of the input viewpoint images.

The heading of each viewpoint image is determined by the agent's path. For any viewpoint $v_{i>0}$, its heading is aligned with the direction of arrival from the previous viewpoint $v_{i-1}$. The initial orientation at the start viewpoint $v_0$ is determined by the episode's start heading.

We handle edge cases where the red arrow falls outside the composite three-image view as follows:

\begin{itemize}

\item At $v_0$: If the first move is backward, we re-orient the initial heading to face $v_1$ directly. After the instruction is generated, we manually prepend a relevant command (e.g., "Turn around, ") to the answer.

\item At $v_{i>0}$: We just discard this trajectory. This is a rare occurrence, affecting only about 2\% of all trajectories.

\end{itemize}

The textual component of our prompt is designed to get six distinct instructions for each trajectory within a single API call. To maximize the diversity among these six instructions, we introduce a strategy called trajectory segmentation annotation.

Instead of prompting for the full trajectory directly, we preprocess it by randomly splitting the trajectory into 1/2/3 continuous subtasks. A single prompt includes three different split schemes, and Gemini is asked to generate two descriptions for each sub-task within each scheme. The final six instructions are then constructed by concatenating the descriptions corresponding to each segmentation. This method efficiently yields six varied instructions by encouraging the model to describe the same trajectory at different levels of granularity.

Furthermore, to avoid constraining Gemini's generative capabilities, our prompt deliberately omits any few-shot examples of R2R human-annotated instructions. The primary constraint we impose is on the length of the description for each sub-task, guiding it to be approximately 10-25 words.

As shown in Fig.\ref{fig:data}, the instructions generated by Gemini contain fewer hallucinations and show enhanced linguistic richness compared to the original dataset. Quantitatively, the average instruction length of our final \textbf{Prevalent\_Gemini\_Aug} dataset increased from 31 words (original Prevalent) to 48 words.

\subsection{Model Architecture}
\label{sec:model}

The workflow of our model begins by encoding the RGBD features of graph nodes into a sequence of node tokens. These tokens are then jointly processed with the encoded instructions by the DPFT module. Finally, a Single Action Prediction (SAP) head scores the output features from the DPFT, selecting the node with the highest score as the next subgoal, and the low-level controller navigates the agent toward it. The remainder of this subsection will focus on detailing the components of our high-level planner: the input encoders, the DPFT module, and the task heads.

\subsubsection{Text Encoder} 
\label{sec:text encoder}
The language instruction $W=\{w_i\}_{i=1}^{N_t}$ is first tokenized and mapped to word embeddings. These embeddings are then summed with positional and type embeddings, along with a novel task embedding that we introduce. This task embedding is conditioned on the instruction's source—R2R (type 1), RxR (type 2), or our Gemini-annotated data (type 3)—allowing the model to handle linguistic variations during our joint pretraining (detailed in Section \ref{sec:train}). The resulting sequence of embeddings is then fed into a multi-layer BERT-style Transformer encoder \cite{devlin2019bert}, which outputs the final contextualized instruction features $T=\{t_i\}_{i=1}^{N_t}$.

\subsubsection{Node Encoder} 
For each node in graph, we first process its panoramic observation $O_t=\{I^{rgb}_k, I^{d}_k\}_{k=1}^{12}$ by passing each RGB and depth image through respective vision backbones to extract unimodal features. These features are then combined with a learnable view angle embedding. After being projected through separate linear layers, these three features are fed into a multi-layer Transformer (panorama encoder) to fuse the panoramic context of the node. The final node token for the current and previously visited nodes is obtained by averaging 12 output vectors. For unvisited nodes, the feature is directly inherited from the corresponding view-specific feature of its parent node. Finally, each node token is summed it with three additional embeddings: a step embedding (encoding the last visit time), a position embedding (encoding the relative pose to the current node), and a task embedding (ID 1 for R2R, 2 for RxR) similar to \ref{sec:text encoder}. All processed node features are then collected into the final graph token sequence $G=\{g_i\}_{i=1}^{N_g}$, which also includes a dedicated token for the STOP action.

\begin{figure}[t]
    \centering
    \includegraphics[width=\columnwidth]{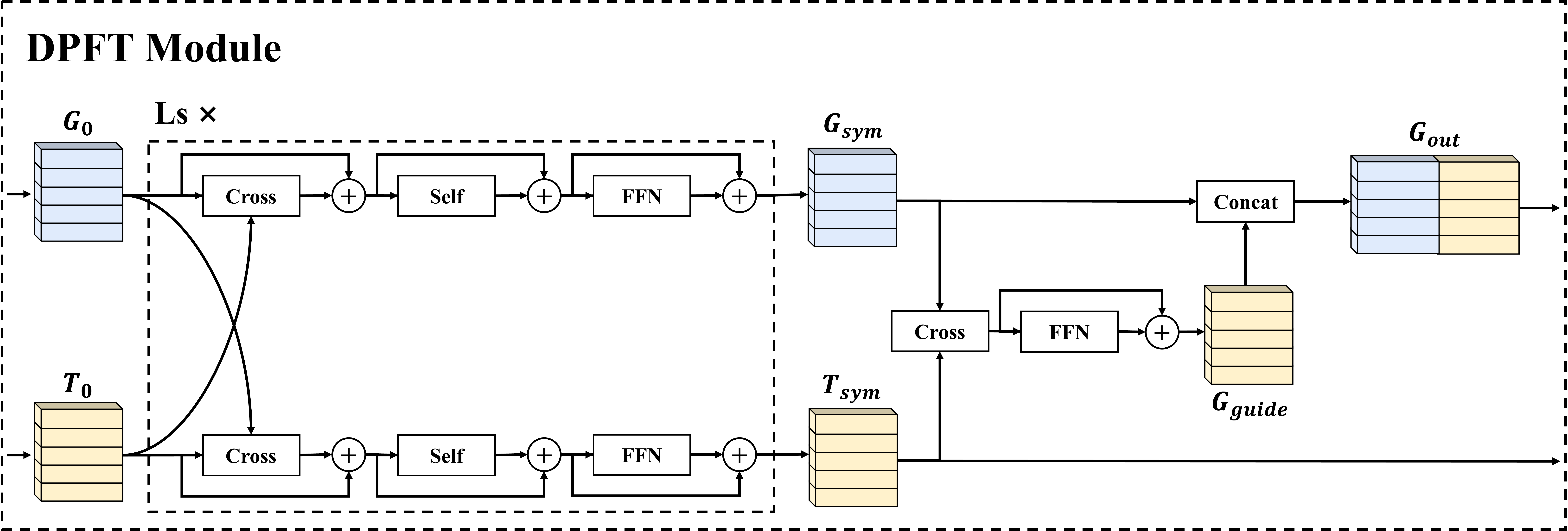}
    \caption{Illustration of the DPFT framework.}
    \label{fig:model}
    \vspace{-15pt}
\end{figure}

\subsubsection{DPFT Module}

As shown in Figure \ref{fig:model}, our DPFT module begins with the Symmetric Cross-Modal Fusion network, consists of $L_s$ Transformer layers that create mutually-aware representations of the instruction and the graph. Let the initial inputs be the text features $T_{0} \in \mathbb{R}^{N_{t} \times d}$ and the graph features $G_{0} \in \mathbb{R}^{N_{g} \times d}$. For each layer $l$ from 0 to $L_s-1$, bidirectional cross-attention modules first generates intermediate representations. The text features attend to the graph, and vice-versa. Unlike LXMERT \cite{tan2019lxmert}, we utilize unshared weights for the two cross-attention modules as a scaling strategy. These intermediate representations are then processed by self-attention and feed-forward network (FFN) to produce the layer's output, $T_{l+1}$ and $G_{l+1}$.

The outputs of this symmetric phase, $T_{sym}=T_{L_s}$, $G_{sym}=G_{L_s}$, are then fed into the Text-Guided Graph Refinement network. This module further refines the graph representation by injecting specific guidance distilled from the text. The graph features $G_{sym}$ query the text features $T_{sym}$ to extract a language guidance vector $G_{guide}$: 
\begin{equation}
G_{{guide}}=\text{CrossAttn}(Q=G_{{sym}},K=T_{{sym}},V=T_{{sym}})
\label{eq:2}
\end{equation}
\noindent This vector, representing the most relevant textual information for each graph node, is processed by an FFN and concatenated with the original graph features to produce the final, guidance-enriched graph representation, $G_{out}$:
\begin{equation}
G_{{out}}=\operatorname{Concat}\left(G_{{sym}}, \operatorname{FFN}\left(G_{ {guide}}\right)\right)
\label{eq:3}
\end{equation}
The final outputs of entire DPFT module are the contextualized text features $T_{sym}$ and the refined graph features $G_{out}$.

\subsubsection{Task Heads}
Our model employs two task heads, both implemented as simple FFNs. The primary Single Action Prediction (SAP) head is responsible for scoring each candidate node in the graph; the auxiliary Masked Language Modeling (MLM) head is used during pretraining for predicting the vocabulary distribution of masked tokens:
\begin{align}
s_{i}&=\mathrm{FFN}_{sap}\left(g_{i}\right) \\
l_{j}&=\mathrm{FFN}_{mlm}\left(t_{j}\right)
\label{eq:4}
\end{align}
\noindent where $g_i$ is the feature vector for $i$-th candidate node extracted from $G_{out}$. Similarly, $t_j$ is the feature vector for $j$-th masked token from $T_{sym}$.

\subsection{Training Paradigm}
\label{sec:train}
As shown in Figure \ref{fig:overview}, our model is trained in a three-stage paradigm to progressively enhance its capabilities. We first obtain a base model through offline joint pretraining on a large corpus of data from both R2R and RxR tasks. This base model is then fine-tuned on each task separately using online DAgger-based SFT and finally polished with online GRPO-based RFT. The three stages are detailed below.

\subsubsection{Offline Joint Pretraining}
To enhance generalization, we pretrain a single base model by jointly utilizing five datasets from R2R and RxR: Prevalent (1M) \cite{hao2020towards}, our Prevalent\_Gemini\_Aug (1M), RxR-Marky (1M) \cite{wang2022marky}, and the smaller original task datasets R2R\_train (14K) \cite{VLN} and RxR\_train (80K) \cite{RXR}. The pretraining curriculum consists of two auxiliary tasks with 1:1 ratio: Single Action Prediction (SAP) and Masked Language Modeling (MLM). For SAP, the model predicts the next action given the instruction and a partial ground-truth trajectory. For MLM, it reconstructs masked tokens in the instruction, conditioned on full ground-truth trajectory. The best-performing checkpoint is selected based on the highest sum of SAP and MLM accuracies on the R2R\_val\_unseen + RxR\_val\_unseen validation sets.

\subsubsection{Online SFT}
To improve the agent's ability to explore and self-correct from its own mistakes, we perform online SFT using the DAgger algorithm \cite{dagger}. At each step, the agent follows the expert action with probability $p$ and a sampled action from its own policy with probability $1-p$. The expert actions are optimal nodes provided by a global planner, following the same setup in ETPNav \cite{an2024etpnav}. We select the final checkpoints for R2R-CE based on the highest Success Rate + SPL, and for RxR-CE based on the highest nDTW + SDTW on their respective val\_unseen sets.

\subsubsection{Online RFT}
For the final stage, we perform closed-loop RFT using GRPO \cite{shao2024deepseekmath}, an efficient RL algorithm. GRPO is a variant of PPO \cite{PPO} that foregoes the need for a critic network, significantly reducing the memory and computational burden of RL training. Instead of learning a value function, GRPO samples a group of $G$ answers for the same prompt and calculates the advantage for each answer relative to the group's average performance. In the context of graph-based VLN, an episode serves the role of a prompt, a trajectory corresponds to an answer, and a high-level action represents a token. This direct correspondence enables the use of a simple and interpretable outcome-based reward for the full episode. We define the reward function as:
\begin{subequations}
\label{eq:reward}
\begin{align}
R_{\mathrm{R} 2 \mathrm{R}-\mathrm{CE}}&=\mathbb{I}\left(d_{\text {final }}<1.5\right)+\mathrm{SPL}-d_{\text {final }} / 6 \\
R_{\mathrm{RxR}-\mathrm{CE}}&=\mathrm{nDTW}+\mathrm{SDTW}+\mathrm{gSPL}-d_{\text {final }} / 6
\end{align}
\end{subequations}
where $\mathbb{I}(\cdot)$ is the indicator function for success, and $d_{\text {final }}$ is the agent's final geodetic distance to the goal. gSPL is a variant of SPL, it uses the ground-truth path length instead of the shortest path length in SPL. This distinction is critical for RxR-CE, as its reference trajectories are often intentionally non-shortest paths. Using SPL would incorrectly incentivize the agent to deviate from the instructions in favor of taking shortcuts. In contrast, gSPL correctly rewards paths that are efficient while still complying with to the given instruction. 

For each episode, we sample a group of $G$ trajectories and calculate their episode rewards $\textbf{r}=\{r_1, ..., r_G\}$ based on Eq. \ref{eq:reward}. These rewards are then normalized and assigned as the advantage value $\hat{A}_{i, t}$ for all high-level steps $t$ within the $i$-th trajectory:
\begin{equation}
\hat{A}_{i, t}=\widetilde{r}_{i}=\frac{r_{i}-\operatorname{mean}(\mathbf{r})}{\operatorname{std}(\mathbf{r})}
\end{equation}

Our final optimization objective for GRPO, which regularizes the policy update with a KL divergence term against the reference policy, is the same as its original form \cite{shao2024deepseekmath}.

\section{Experiment}

\begin{table*}[t]
\centering
\caption{Experimental results on R2R-CE and RxR-CE datasets}
\label{tab:result}
\begin{tabular}{@{}lccccccccccccc@{}}
\toprule
& \multicolumn{8}{c}{R2R-CE} & \multicolumn{5}{c}{RxR-CE} \\
\cmidrule(lr){2-9} \cmidrule(lr){10-14}
& \multicolumn{4}{c}{Val Unseen} & \multicolumn{4}{c}{Test Unseen} & \multicolumn{5}{c}{Val Unseen} \\
\cmidrule(lr){2-5} \cmidrule(lr){6-9} \cmidrule(lr){10-14}
Methods & NE$\downarrow$ & OSR$\uparrow$ & \textbf{SR}$\uparrow$ & \textbf{SPL}$\uparrow$ & NE$\downarrow$ & OSR$\uparrow$ & \textbf{SR}$\uparrow$ & \textbf{SPL}$\uparrow$ & NE$\downarrow$ & SR$\uparrow$ & SPL$\uparrow$ & \textbf{nDTW}$\uparrow$ & \textbf{SDTW}$\uparrow$\\
\midrule
VLN$\circlearrowright$BERT \cite{hong2022bridging}     & 5.74 & 53 & 44 & 39  & 5.89 & 51 & 42 & 36   & 8.98 & 27.08 & 22.65 & 46.71 & { - } \\
GridMM \cite{wang2023gridmm}       & 5.11 & 61 & 49 & 41  & 5.64 & 56 & 46 & 39   & { - }   & { - }    & { - }    & { - } & { - }  \\
ScaleVLN \cite{wang2023scalingvln}      & 4.80 & { - }  & 55 & 51  & 5.11 & { - }  & 55 & 50   & { - }   & { - }    & { - }    & { - }  & { - }  \\
BEVBert \cite{an2022bevbert}     & 4.57 & 67 & 59 & 50  & 4.70 & 67 & 59 & 50   & { - }   & { - }    & { - }    & { - } & { - }    \\
ETPNav \cite{an2024etpnav}       & 4.71 & 65 & 57 & 49  & 5.12 & 63 & 55 & 48   & 5.64 & 54.79 & 44.89 & 61.90 & 45.33 \\
HNR \cite{hnr}          & 4.42 & 67 & 61 & 51  & 4.81 & 67 & 58 & 50  & 5.51 & 56.39 & 46.73 & 63.56 & 47.24 \\
G3D-LF \cite{g3d}       & 4.53 & 68 & 61 & 52  & 4.78 & 68 & 58 & 51  & { - }   & { - }    & { - }    & { - } & { - }    \\
\midrule
Ours-DAgger    & \underline{4.11} &  \underline{69} &  \underline{63} &  \underline{54}  & { - }  & { - } & { - }   & { - }   &  \underline{5.42}    &  \underline{58.26}    &  \underline{48.19}  &  \underline{63.78} &  \underline{48.53}  \\
Ours-GRPO     & \textbf{3.94} & \textbf{72} & \textbf{65} & \textbf{56}  & \textbf{4.19} & \textbf{69} & \textbf{64} & \textbf{54} & \textbf{5.22}   & \textbf{59.92}    & \textbf{48.97}    & \textbf{65.31}  & \textbf{50.41} \\
\bottomrule
\vspace{-15pt}
\end{tabular}
\end{table*}

\subsection{Experimental Setup}

\subsubsection{Datasets}
Our experiments are conducted on two standard VLN-CE benchmarks, \textbf{R2R-CE} and \textbf{RxR-CE}, which adapt the original R2R \cite{VLN} and RxR \cite{RXR} datasets for navigation in the continuous Habitat \cite{savva2019habitat} environment. The two datasets present distinct challenges in language complexity, path structure, and physical constraints. R2R-CE features concise, English-only instructions with an average length of 32 words, and its reference paths are always the shortest path to the destination, averaging 9.89m in length. In contrast, RxR-CE provides highly descriptive, multilingual instructions (English, Hindi, and Telugu) that average 120 words. Its paths are longer (15.23m on average) and often follow non-shortest routes, demanding the agent to strictly follow the detailed instructions. The benchmarks also differ in their agent simulation parameters: the agent in R2R-CE has a smaller chassis radius of 0.10m and is permitted to slide along obstacles, whereas the RxR-CE agent has a larger 0.18m radius and is stopped by collisions.

\subsubsection{Evaluation Metrics}
We adopt a standard suite of navigation metrics: Navigation Error (NE), Success Rate (SR), Oracle SR (OSR), Success weighted by Path Length (SPL), normalized Dynamic Time Warping (nDTW), and Success weighted by nDTW (SDTW).  For the R2R-CE benchmark, we prioritize \textbf{SR and SPL} to evaluate success and path efficiency. For the RxR-CE benchmark, where fidelity to the reference path is more critical, we focus on \textbf{nDTW and SDTW}.

\subsubsection{Implementation Details}
For our visual encoders, we use a ViT-B/32 \cite{dosovitskiy2020image} pretrained on CLIP \cite{radford2021clip} for RGB images and a ResNet-50 \cite{he2016deep} pretrained on point-goal navigation \cite{wijmans2019ddppo} for depth images. For the text encoder, we adopt the 12-layer RoBERTa \cite{liu2019roberta} architecture and initialize the model with its pretrained weights. The number of layers for the panorama encoder and the Symmetric Cross-Modal Fusion network are set to 2 and 4, respectively. The hidden dimension $d$ for the entire model is set to 768. During the pretraining and online SFT stages, all model parameters are trained except for the frozen visual backbones. In the final online RFT stage, we exclusively fine-tuning the DPFT module and the SAP head. 

For the online training stages, we partition the dataset into two disjoint sets: 90\% of the data is allocated for Online SFT, while the remaining 10\% is reserved for Online RFT.

To further improve RFT performance, we implement several specific configurations: (1) dropout is enabled for frozen layers; (2) dropout is also active during the sampling phase; and (3) the GRPO update iteration number $\mu$ is set to 1. The impact of these configurations is analyzed in Section \ref{sec:GRPO_ablation}. Additional GRPO hyperparameters include a clip range $\epsilon=0.2$, a KL loss weight $\beta=0.04$, and a group size $G=8$.

\subsection{Comparison With State-of-the-Art Methods}
\subsubsection{R2R-CE}
The left part of Table \ref{tab:result} presents a comparison of our ETP-R1 against current state-of-the-art approaches on the R2R-CE benchmark. The results indicate that our approach outperforms all previous methods across all splits on the key metrics of NE, OSR, SR, and SPL. Notably, our model trained only with online SFT on 90\% data, denoted as Ours-DAgger, already establishes a new SOTA. On the Val Unseen split, it reduces NE by at least 0.3m against prior methods, while improving SR and SPL by 2\% respectively compared to the second-best method G3D-LF \cite{g3d}. This strong performance highlights the effectiveness of our network architecture and the rich, unified pretraining dataset in producing a powerful base model.

\begin{figure*}[t]
    \centering
    \includegraphics[width=\textwidth]{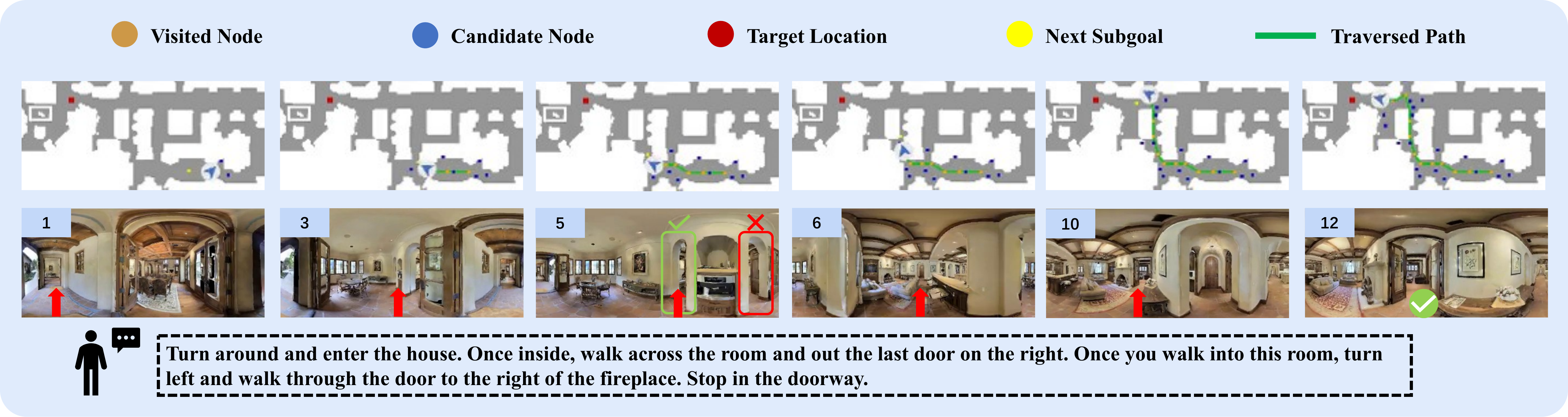}
    \vspace{-15pt}
    \caption{This episode from R2R-CE Val Unseen showcases our model's ability to follow complex instructions. The agent successfully handles a long trajectory by (1) following the backward command, (2) selecting the correct door from two ambiguous options, and (3) following subsequent directional commands to reach the destination.}
    \label{fig:case study}
    \vspace{-10pt}
\end{figure*}

Further applying GRPO-based online RFT on the remaining 10\% data yields additional performance gains. By rewarding trajectories that stop closer to the destination, achieve success, and follow more efficient paths, the final model, Ours-GRPO, further reduces NE by 0.18m and increases both SR and SPL by another 2\% on the Val Unseen split. Figure \ref{fig:case study} illustrates a case from R2R Val Unseen, showcasing the strong instruction-following ability of Ours-GRPO. Ultimately, on the Test Unseen split, Ours-GRPO surpasses the previous state-of-the-art G3D-LF \cite{g3d} by a large margin, lowering the NE by 0.59m while achieving a 6\% higher SR and a 3\% higher SPL. This demonstrates the robust generalization capability of our final model.

\subsubsection{RxR-CE}
The right part of Table \ref{tab:result} shows our results on the RxR-CE Val Unseen. Similarly, the Ours-DAgger model without RFT already outperforms the prior SOTA method, HNR \cite{hnr}. The Ours-GRPO model further improves the performance, ultimately surpassing HNR on all metrics. It reduces NE by 0.29m, and increases SR, SPL, nDTW, and SDTW by 3.53\%, 2.24\%, 1.75, and 3.17, respectively, setting a new state-of-the-art. The experimental results confirm the positively impact of cross-dataset pretraining and GRPO fine-tuning  to the final RxR-CE performance, even without expanding the RxR task-specific pretraining data.

\subsection{Ablation Study}
We conduct extensive experiments to validate the key design choices of our ETP-R1 model concerning the pretraining dataset, model architecture, and GRPO configurations. All results are reported using either the R2R validation metrics during pretraining or on the R2R-CE Val Unseen split.

\begin{figure*}[t]
    \centering
    \includegraphics[width=0.9\textwidth]{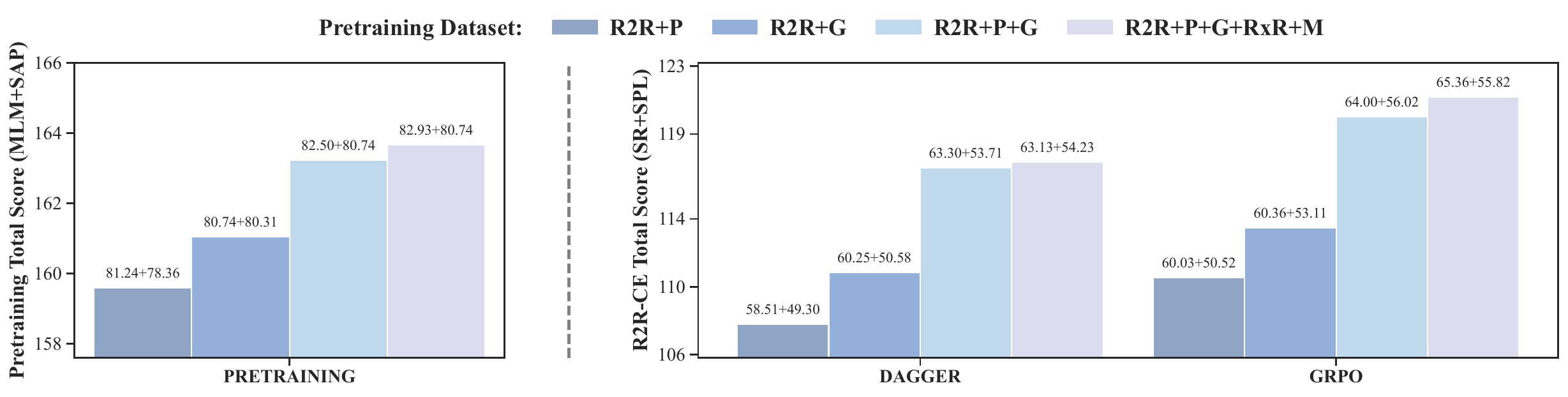}
    \caption{Ablation on Pretraining Data Composition} 
    \label{fig:dataset_ablation}
    \vspace{-20pt}
\end{figure*}

\subsubsection{Ablation on Pretraining Data Composition}
\label{sec:pretrain ablation}
We conduct a detailed analysis of different pretraining dataset compositions, evaluating their impact on both the pretraining phase and the subsequent performance during online fine-tuning. As shown in Figure \ref{fig:dataset_ablation}, we measure pretraining performance as the sum of MLM and SAP accuracies on R2R\_val\_unseen, and online performance as the sum of SR and SPL on the R2R-CE Val Unseen split. The datasets used are abbreviated as follows: R2R (R2R\_train), RxR (RxR\_train), P (Prevalent), G (Prevalent\_Gemini\_Aug), and M (RxR-Marky).

A key initial observation is the strong positive correlation between pretraining scores and final online performance across all four setups, indicating that a stronger base model consistently leads to a better final agent. Comparing R2R+G to R2R+P highlights the superiority of our Gemini-annotated data. R2R+G improves the pretraining score by 1.45\% and the final performance by approximately 3\%. Notably, while R2R+P achieves a higher MLM accuracy due to the in-domain nature of the speaker model used for P, its monotonous and often hallucinatory instructions lead to a significantly lower SAP accuracy, demonstrating the importance of high-quality language for learning robust action policies. Then, the comparison between R2R+P+G and the previous settings shows that scaling up the dataset provides substantial benefits. Stacking both synthetic datasets improves the pretraining score by 2.19\% and the final performance by a significant 6.55\% over R2R+G, which suggests that VLN pretraining is still data-hungry. Finally, we investigate the effect of cross-task data by comparing R2R+P+G+RxR+M with R2R+P+G. The inclusion of RxR-related data further improves the pretraining and final R2R performance by 0.43\% and 1.16\%, respectively, confirming that the model learns transferable knowledge from related tasks. However, the magnitude of this improvement is smaller than that gained from adding more in-domain data.

\begin{table}[t]
\centering
\caption{Ablation study on model architecture components. We report the pretraining scores as results.}
\label{tab:model}
\begin{tabular}{@{}lccc@{}}
\toprule
Methods & MLM$\uparrow$ & SAP$\uparrow$  & \textbf{Total}$\uparrow$\\
\midrule
\textbf{Full Model} & {82.93} & {80.74} &\textbf{163.67}\\
\hspace*{1em}- Text-Guided& 83.10 & 79.72 &162.82\\
\hspace*{1em}- Task Emb& 82.21 & 80.87 &163.08\\
\bottomrule
\vspace{-20pt}
\end{tabular}
\end{table}

\subsubsection{Ablation on Architectural Components}
We perform an ablation study to verify the effectiveness of two key components in our model architecture: the Text-Guided Graph Refinement network and the task embedding. Based on the strong correlation between pretraining and final performance established in the Section \ref{sec:pretrain ablation}, we use the pretraining score on R2R\_val\_unseen as the primary metric. The results are summarized in Table \ref{tab:model}. First, removing the Text-Guided Graph Refinement module from the DPFT (- Text-Guided), which results in a purely symmetric fusion architecture, causes a 0.85\% drop in the total score. Second, removing the task embedding from both text and node features (- Task Emb) leads to a 0.59\% performance decrease. These results confirm that both proposed components are beneficial. We attribute the less pronounced impact of these architectural changes, compared to that of the dataset composition, to our model's data-hungry nature, where the benefits of architectural refinements may not be fully shown.

\begin{table}[t]
\centering
\caption{Ablation study on GRPO settings.}
\label{tab:grpo}
\begin{tabular}{@{}lccc@{}}
\toprule
Methods & SR$\uparrow$ & SPL$\uparrow$ &\textbf{Total}$\uparrow$\\
\midrule
\textbf{Ours-GRPO} & 65.36 & 55.82 &\textbf{121.18} \\ 
\hspace*{1em} - Sample Dropout & 63.78 & 55.10 &118.88\\ 
\hspace*{1em} - Frozen Dropout & 64.17 & 54.96 &119.13\\
\hspace*{1em} + Temperature Scaling & 63.46 & 53.80 &117.26\\
\hspace*{1em} + Multi-Epoch Update ($\mu$=2) & 64.76 & 54.25 &119.01\\
\bottomrule
\vspace{-20pt}
\end{tabular}
\end{table}

\subsubsection{Ablation on GRPO settings}
\label{sec:GRPO_ablation}
We conduct ablation studies on four different GRPO setups to demonstrate the rationale behind our final configuration, with results on the R2R-CE Val Unseen split summarized in Table \ref{tab:grpo}. The first two variants investigate the role of dropout during RFT. Disabling dropout during sampling phase (- Sample Dropout) or for frozen layers (- Frozen Dropout) both result in a performance drop of approximately 2\% in the total score. This indicates that maintaining dropout is crucial for GRPO's effectiveness on a well-trained model. We hypothesize that dropout introduces beneficial stochasticity, particularly during sampling, which leads to the collection of more diverse trajectories. This enhanced exploration helps prevent the policy from collapsing into a suboptimal local minimum.

Next, we explore an alternative method for encouraging sampling diversity by applying temperature scaling to the action logits (+ Temperature Scaling), with T decaying from 2 to 1. This approach performs significantly worse, with results comparable to our pre-RFT Ours-DAgger model. A possible explanation is that temperature scaling incentivizes the model to produce even sharper logits to counteract the softmax flattening and secure high rewards—a process of reinforcing confidence that does not necessarily involve learning new navigational knowledge. Finally, we test a variant where each sampled batch is used for two consecutive parameter updates (+ Multi-Epoch Update ($\mu$=2)). This also degrades performance by about 2\%, suggesting that for the VLN task, GRPO performs optimally in a strictly on-policy manner (one update per data batch).

\section{Conclusion}
In summary, this paper introduces ETP-R1, a novel graph-based VLN framework that is the first to apply closed-loop RFT. To build a strong base model, we first enhance the pretraining data by Gemini API to generate more instructions and perform joint pretraining across R2R and RxR datasets. Ablation studies indicate that our Gemini-annotated data is superior to its predecessor and that scaling the dataset size yields substantial performance gains. Building on this, we introduce an efficient GRPO-based online RFT stage, which further boosts performance, demonstrating the effectiveness of our training paradigm. Our ablation studies highlight key factors for successful RFT, such as the crucial role of dropout. As a result of these contributions, our final model achieves new state-of-the-art performance across all major metrics on both the R2R-CE and RxR-CE benchmarks.



\bibliography{root}

\bibliographystyle{ieeetr}

\end{document}